\documentclass[letterpaper, 10 pt, conference]{ieeeconf}  

\IEEEoverridecommandlockouts                              

\usepackage{algorithm}
\usepackage{algpseudocode}
\usepackage{hyperref}
\usepackage{graphicx} 
\usepackage{caption,subcaption}

\title{\LARGE \bf
A Virtual Reality Teleoperation Interface for \\Industrial Robot Manipulators
}

\author{Eric Rosen$^{1}$ and Devesh K. Jha$^{2}$
\thanks{$^{1}$Eric Rosen is with Department of Computer Science, Brown University, Providence, RI.
        {\tt\small eric\_rosen@brown.edu}}%
\thanks{$^{2}$Devesh K. Jha is with Mitsubishi Electric Research Labs, Cambridge, MA 
        {\tt\small jha@merl.com}}%
}

\begin{document}

\maketitle
\thispagestyle{empty}
\pagestyle{empty}

\begin{abstract}
We address the problem of teleoperating an industrial robot manipulator via a commercially available Virtual Reality (VR) interface. Previous works on VR teleoperation for robot manipulators focus primarily on collaborative or research robot platforms (whose dynamics and constraints differ from industrial robot arms), or only address tasks where the robot's dynamics are not as important (e.g: pick and place tasks). We investigate the usage of commercially available VR interfaces for effectively teleoeprating industrial robot manipulators in a variety of contact-rich manipulation tasks. We find that applying standard practices for VR control of robot arms is challenging for industrial platforms because torque and velocity control is not exposed, and position control is mediated through a black-box controller. To mitigate these problems, we propose a simplified filtering approach to process command signals to enable operators to effectively teleoperate industrial robot arms with VR interfaces in dexterous manipulation tasks. We hope our findings will help robot practitioners implement and setup effective VR teleoperation interfaces for robot manipulators. The proposed method is demonstrated on a variety of contact-rich manipulation tasks which can also involve very precise movement of the robot during execution (videos can be found at \url{https://www.youtube.com/watch?v=OhkCB9mOaBc}).

\end{abstract}

\section{INTRODUCTION}
\begin{figure*}
    \centering
    \begin{subfigure}{0.47\textwidth}
    \centering
        \includegraphics[width=0.9\linewidth]{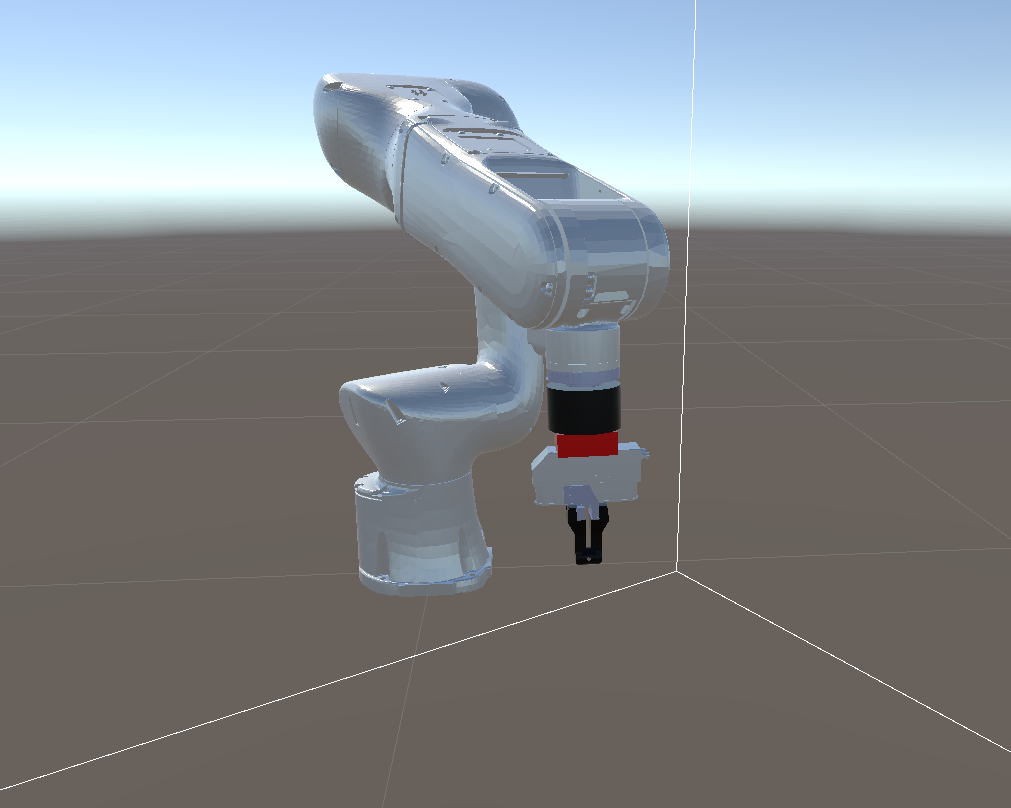}
        \caption{Virtual reality rendering of the MELFA Assista robot}
        \label{fig:vr_render}
    \end{subfigure}%
    \begin{subfigure}{0.47\textwidth}
    \centering
        \includegraphics[width=0.85\linewidth]{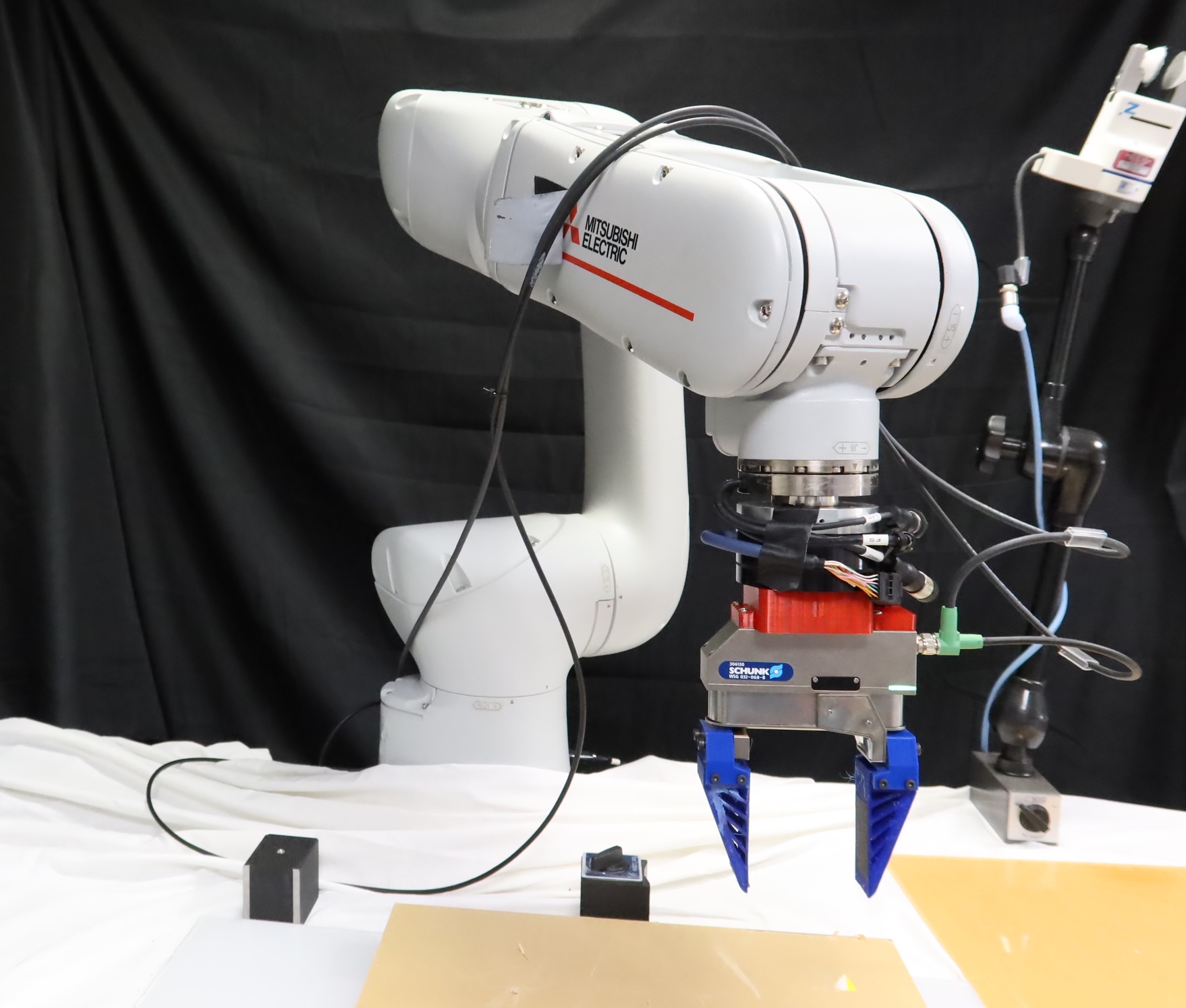}
        \caption{Actual Image of the MELFA Assista robot used in the study}
        \label{fig:assista_image}
    \end{subfigure}
    \caption{VR rendering and the actual robot image used in this study.}
    \label{fig:vr_and_actual_image}
\end{figure*}

Robots have been steadily increasing their presence in our daily lives where they can be used to create and provide more assistance for various collaborative tasks. It is envisioned that the next-generation industrial robotic systems will have humans and robots working next to each where robots can easily learn complex tasks from human experts as part of the collaborative system. However, creating robotic systems which can easily  collaborate while being versatile is extremely challenging. One of the key challenges is to design interfaces which be used intuitively for a large variety of tasks for collaborative human-robot systems. Such interfaces are very important for designing collaborative systems so that a human can easily demonstrate complex manipulation tasks to the robot. In this paper, we consider teleoperation of industrial robots using virtual reality interfaces and demonstrate the efficiency of the system over a wide variety of manipulation tasks. Figure~\ref{fig:vr_and_actual_image} shows a rendering of the manipulator arm in the virtual game environment as well as an actual image of the robot used in this study.

\begin{figure}
    \centering
        \includegraphics[width=\linewidth]{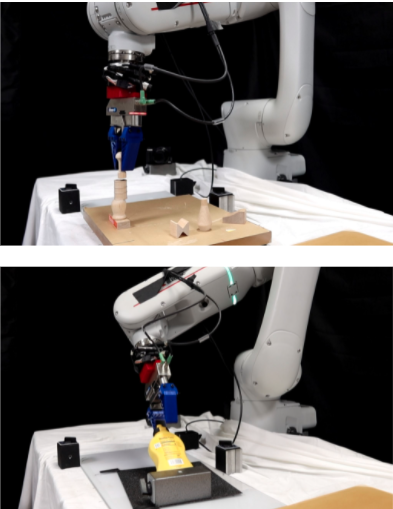}
    \caption{Example images from two of the manipulation tasks we succesfully performed with our VR teleoperation interface. Top: stacking Bandu pieces. Bottom: Flipping a bottle upwards using extrinsic contacts.}
    \label{fig:front}
\end{figure}

Teleoperation is the problem and set of solutions associated with remotely controlling a semi-autonomous robot platform. Traditionally, teleoperation is performed by system experts because standard interfaces like keyboards, mice, and teach pendents require advanced training to effectively use. More recently, commercially available Virtual Reality (VR) hardware have gained tractions as viable interfaces for robot teleopertaion because the spatially-tracked hand controllers make it easier for non-experts to command high DoF arms \cite{whitney2019comparing}. Most research into VR teleoperation for robots has done work on research robot platforms. We address the problem of applying these techniques to industrial robot arm platforms. 

In general, the industrial robotic arms with pure position control have several characteristics which make VR teleoperation very challenging for these manipulators when compared to research robotic platforms. Firstly, industrial robot arms focus primarily on precision, and therefore often only have high-precision position control scheme exposed, whereas many research robot platforms have access to velocity or torque control. Secondly, industrial robot arms tend to be heavier due to the mechanical construction compared to cobots or research platforms, and so the higher amounts of inertia introduce a higher amount of latency between control signals from the operator to commanded movements on the robot platform. We find that these two important differences make it extremely challenging to take existing practices for VR teleoperation developed in research laboratories and effectively deploy them on industrial robot arms to perform dexterous manipulation tasks (Figure \ref{fig:front}).

To overcome these issues, we propose some key system implementations that improve teleoperation efficacy. We propose a signal-processing algorithm for interpolating user-specified position controls for a robot arm that prevents the robot from faulting even when the robot's controller dynamics are not known or adjustable. We also use a hybrid force-position control scheme where operators command both the robot's end-effector position (via spatially-tracked hand controllers) while using the robot in impedance control mode to help compensate for discrepancies due to latency as well as provide some compliance during manipulation tasks.

Our paper is structured as follows: we first describe existing practices and approaches for VR teleoperation of robot manipulators, and describe in detail the challenges we found when applying them to an industrial robot arm. Then, we describe in detail our proposed set of system practices for improving the efficacy of VR teleoperation for industrial robot arms. Finally, we conclude with an experimental demonstration of how our system changes enable operators to teleoperate an industrial robot arm with VR to perform dexterous manipulation tasks like stacking, peg-in-hole insertion and various other challenging manipulation tasks. 

We believe our findings and contributions will be useful to practitioners who aim to deploy existing VR teleoperation research to industrial robot platforms.

\section{RELATED WORK}
Virtual, augmented, and mixed reality devices have recently garnered interest within the robotics community as powerful visual interfaces for facilitating human-robot interaction \cite{chang2022virtual}.
Virtual Reality interfaces for teleoperating robots have empirically improved the performance of novice operators in metrics such as increased task success and lower cognitive workload \cite{whitney2019comparing,barentine2021vr,wonsick2020systematic}. In this section, we discuss different VR systems and interfaces that have been developed for robot teleoperation in both research and industrial contexts.

\cite{lipton2017baxter} proposed a VR system to teleoperate a Baxter robot using a homonuculus view, which embeds the user's view in a control room that is located inside the ``head'' of the robot. This decouples the view of the user from the view of the robot, which makes it easy to control the robot arm based on the distance between the user's hand and the robot's virtual arm markers. \cite{kot2018application} created a virtual reality operator station for the Military mobile robot TAROS which enabled users to command a virtual arm of the robot while accounting for obstacle avoidance. 

Most similar to our work are those that investigate developing VR teleoperation interfaces for industrial robot arms. \cite{solanes2020teleoperation} developed an Augmented Reality (AR) interface for visualizing the state of an industrial robot arm and used a gamepad to control the robot arm. \cite{togias2021virtual} developed a VR interface for increasing the flexibility of existing industrial robot workspaces by enabling end-users to specify Cartesian waypoints when programming motion plans. \cite{kebria2019robust} developed a robust adapative control scheme for teleoperating robots amongst delays, which is especially important when latency is high between the end-user and robot platform.

\section{VIRTUAL REALITY TELEOPERATION FOR INDUSTRIAL MANIPULATORS}
In this section, we first motivate why we think using VR teleoperation is difficult for typical industrial manipulators. This discussion is followed by our proposed method for designing the VR teleoperation system.
\subsection{Why is VR Teleoperation difficult for Industrial Robots?}
We investigate developing an VR interface for controlling an industrial robot manipulator, specifically the MELFA RV-5AS-D Assita robot platform (see Figure~\ref{fig:vr_and_actual_image}). The Assista manipulator is a 6 DoF manipulator equipped with a single DoF parallel-jaw gripper. The robot is equipped with Mitsubishi Electric F/T sensor $1$F-FS$001$-W$200$ at the wrist (see Figure~\ref{fig:vr_and_actual_image}). This sensor is used to implement an indirect force controller using the stock stiffness controller of the robot~\cite{jha2022design, jha2022generalizable}. The Assista is an industrial collaborative manipulator arm with pose repeatability of $\pm 0.03$mm.

\begin{algorithm}
\caption{Filter for VR Teleoperation}\label{alg:cap}
\begin{algorithmic}
\Require Delta end-effector pose $\Delta$p, control frequency $f$', max position speed $s_{p}$, max orientation speed $s_{o}$, noise position threshold $n_{p}$, noise orientation threshold $n_{o}$

$\Delta$$p^{p}$, $\Delta$$p^{o}$ = $\Delta$p \Comment{Separate delta position and orientation}

\If{norm($\Delta$$p^{p}$) $> n_{p}$} \Comment{Delta position above noise threshold}
    \State $\Delta$p$_{filtered}^{p}$ = $\Delta$$p^{p}$/ norm($\Delta$$p^{p}$)
    \State $\Delta$p$_{filtered}^{p}$ *= $s_{p}$ 
\Else \Comment{Delta position below noise threshold}
    \State$\Delta$p$_{filtered}^{p}$ = [0,0,0]
\EndIf

\If{norm($\Delta$$p^{o}$) $> n_{o}$} \Comment{Delta orientation above noise threshold}
    \State $\Delta$p$_{filtered}^{o}$ = $\Delta$$p^{o}$/ norm($\Delta$$p^{o}$)
    \State $\Delta$p$_{filtered}^{o}$ *= $s_{o}$ 
\Else \Comment{Delta orientation below noise threshold}
    \State$\Delta$p$_{filtered}^{o}$ = [0,0,0]
\EndIf

\Return linspace([$\Delta$p$_{filtered}^{p}$,$\Delta$p$_{filtered}^{o}$], $\frac{1}{f'}$)

\end{algorithmic}
\label{algo}
\end{algorithm}

In general, industrial manipulator arms focus on very high precision during operation. Consequently, depending on their pay-load capacity, they tend to be bulky as they use very powerful servo motors during movement. To allow high precision, the robots have a well-tuned control method for the movement of the joints which is not exposed to an end-user. Furthermore, these robots, in general, do not provide access to velocity control methods. Instead, there is a high-frequency, continuous trajectory control method exposed for smooth and continuous movement of the robot. Note that this controller can be used to design a psuedo-velocity controller using a pre-defined position trajectory. An important point to note here is that the high inertia of the industrial manipulators introduce significant latency in following the teleoperation commands through a virtual reality setup.

Another important aspect of these robots is that to allow safe operation, controllers of these robots have a fault system which stops the operation of the robot if a certain command sent by an end-user results in saturation of motor commands. To ensure safe operation, this system is not exposed to the end-user and thus, it acts as a black-box system. For the high-frequency position control, this implies that we can only send very small incremental commands such that the robot operates in the feasible range of the motor controllers. This feasible incremental pose command depends on the on the current pose of the robot. To ensure operation of the robot without incurring faults, we need to ensure that we operate within the unknown, feasible set of commands for robot movement (when using the high-frequency position control). This requirement imposes several challenges regarding how the VR commands should be sent to the robot controller so that we achieve a smooth, error-free movement of the robot while following the human teleoperation commands with sufficient accuracy. 

\begin{figure*}
    \centering
    \begin{subfigure}{\textwidth}
    \centering
        \includegraphics[width=\linewidth]{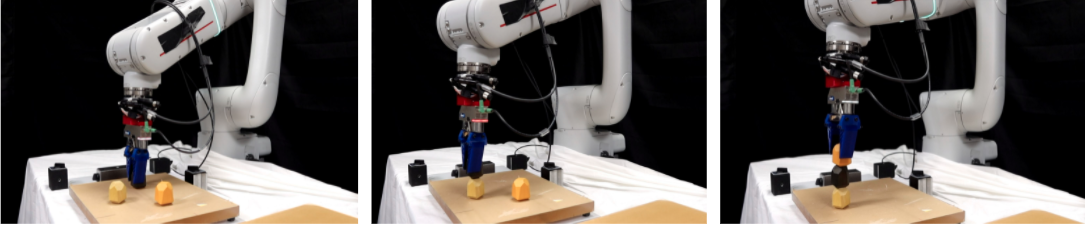}
        \caption{Block stacking (irregular)}
        \label{fig:irregular}
    \end{subfigure}
     \hfil
    \begin{subfigure}{\textwidth}
    \centering
        \includegraphics[width=\linewidth]{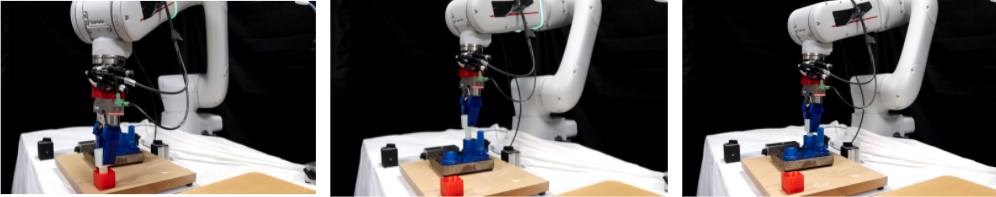}
        \caption{Peg in hole}
        \label{fig:peg-in-hole}
    \end{subfigure}
     \hfil
    \begin{subfigure}{\textwidth}
    \centering
        \includegraphics[width=\linewidth]{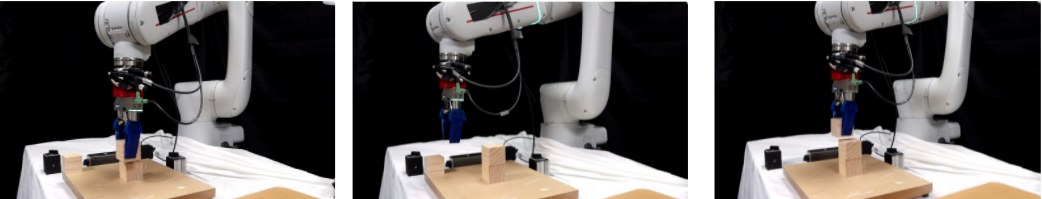}
        \caption{Block stacking}
        \label{fig:block-stacking}
    \end{subfigure}
     \hfil
    \begin{subfigure}{\textwidth}
    \centering
        \includegraphics[width=\linewidth]{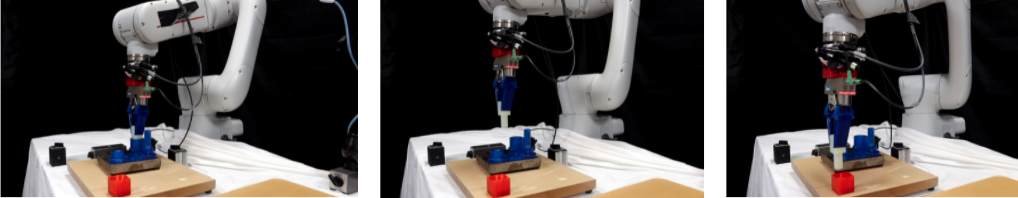}
        \caption{Peg in hole unplugging}
        \label{fig:unplugging}
    \end{subfigure}
    \caption{Sample images of four of the manipulation tasks performed in our experiment using our VR teleoperation interface.}
    \label{fig:tasks1}
\end{figure*}

The above characteristics of a high-precision, position-controlled industrial manipulator arms make it difficult to use a teleoperation system designed for a research robotic system. Consequently, these characteristics require that we filter the signals received from tracking the VR controllers appropriately so that we can achieve low latency while ensuring feasible movement of the manipulator arm. Furthermore, to allow safety during contact interactions, we use the default impedance controller for the robot to allow compliance during contact interactions. One can design a complex compliance controller as shown in our previous work~\cite{jha2022design} but this is out of scope of the current paper.
\subsection{Proposed System}
For controlling the robot arm with a VR headset, we use the HTC Vive Pro. To communicate between the robot and the VR headset, we use ROS\# to generate a virtual environment in Unity3D and create a websockets server to facilitate networking between the ROS and Unity computing nodes. We generate a 3D model of the robot arm in Unity based on the robot's URDF, and update the pose of the model based on the joint states of the real robot. For sending position commands to the robot, we use a similar scheme to \cite{rosen2019communicating}, where a duplicate virtual model of the robot's end effector is instantiated into the scene, and the operator is able to move their 3D spatially-tracked hand-controller to apply relative pose transformations to the copy. Based on the difference between the copy and the robot's end effector transform, a delta end-effector pose $\Delta$p is computed and send as a control command to the robot.

\begin{figure*}
    \centering
    \begin{subfigure}{\textwidth}
    \centering
        \includegraphics[width=\linewidth]{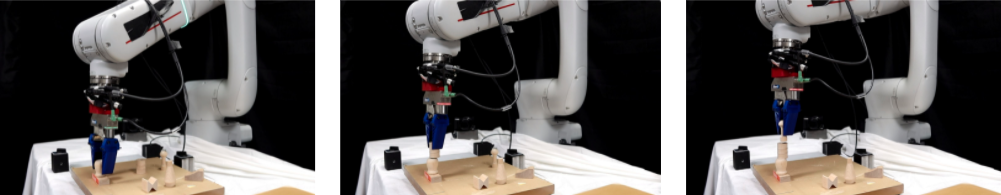}
        \caption{Bandu}
        \label{fig:bandu}
    \end{subfigure}%
    \hfil
    \begin{subfigure}{\textwidth}
    \centering
        \includegraphics[width=\linewidth]{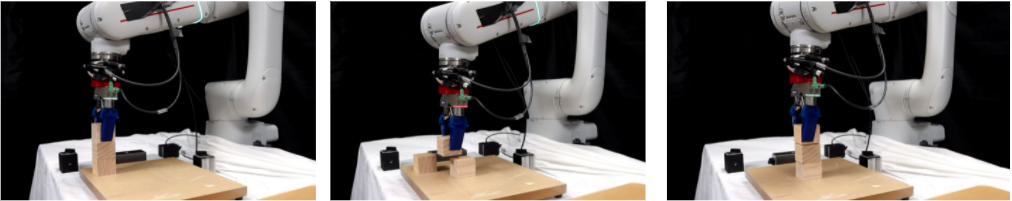}
        \caption{Block destacking}
        \label{fig:unstacking}
    \end{subfigure}
     \hfil
    \begin{subfigure}{\textwidth}
    \centering
        \includegraphics[width=\linewidth]{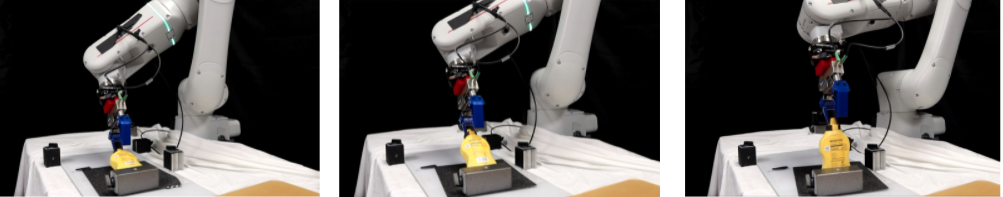}
        \caption{Object flipping}
        \label{fig:flipping}
    \end{subfigure}
     \hfil
    \begin{subfigure}{\textwidth}
    \centering
        \includegraphics[width=\linewidth]{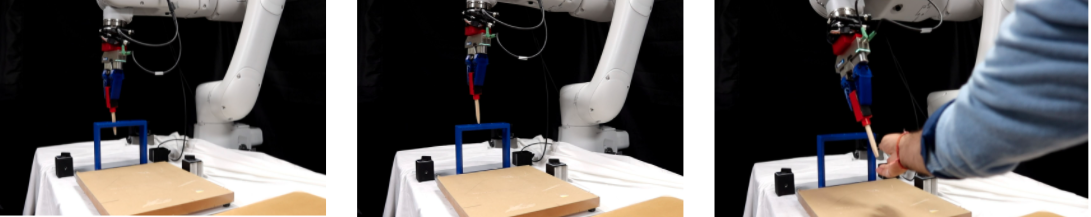}
        \caption{Handover}
        \label{fig:handover}
    \end{subfigure}
    \caption{Sample images of four of the manipulation tasks performed in our experiment using our VR teleoperation interface.}
    \label{fig:tasks2}
\end{figure*}

Given a desired delta end-effector pose $\Delta$p, we can directly use a position-based controller to command the robot arm to the desired configuration. Directly controlling the robot arm via raw position control has been done with relative success in a variety of works using non-industrial robot arms \cite{rosen2019communicating, barentine2021vr}. For these robot platforms, a fixed max speed specification can be set and used to enable the robot to eventually attain any specified $\Delta$p after a finite amount of time. However, for industrial robot platforms like the Assista manipulator, a black-box controller is used to compute sufficient velocities to attain the $\Delta$p at a fixed frequency within a certain accuracy threshold. Due to the robot's built-in fault system, if the commanded $\Delta$p requires a sufficiently high torque to be executed, the robot will fault out, which is problematic for effective teleoperation. This problem is exacerbated since the requisite torques to minimally satisfy the command are a function of the robot's current joint configuration. This means that if a single maximal speed is chosen as a cap to limit how much the user can control the robot's position using the teleoperation system, there will be configurations where the robot motion is limited significantly more than what is necessary.

\begin{figure*}[htb]
    \centering 
\begin{subfigure}{0.25\textwidth}
  \includegraphics[width=\linewidth]{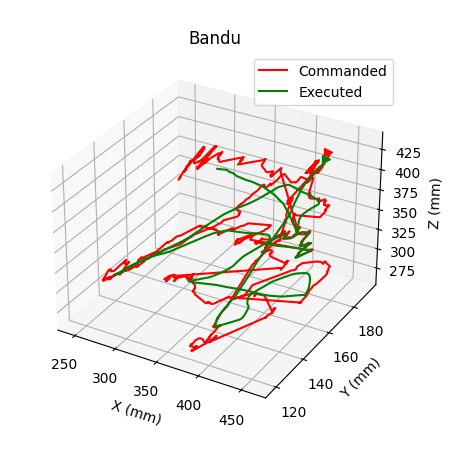}
  \caption{Bandu}
  \label{fig:1}
\end{subfigure}\hfil 
\begin{subfigure}{0.25\textwidth}
  \includegraphics[width=\linewidth]{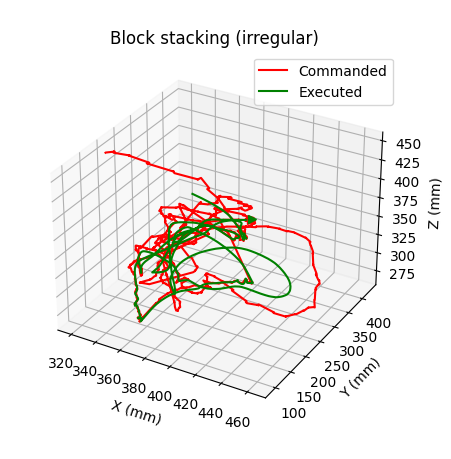}
  \caption{Block destacking (irregular)}
  \label{fig:2}
\end{subfigure}\hfil 
\begin{subfigure}{0.25\textwidth}
  \includegraphics[width=\linewidth]{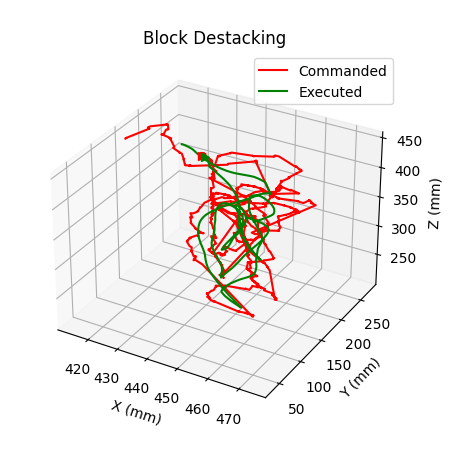}
  \caption{Block destacking}
  \label{fig:3}
\end{subfigure}\hfil
\begin{subfigure}{0.25\textwidth}
  \includegraphics[width=\linewidth]{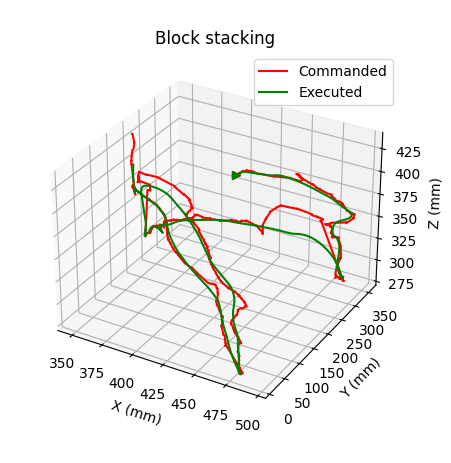}
  \caption{Block stacking}
  \label{fig:4}
\end{subfigure}

\medskip
\begin{subfigure}{0.25\textwidth}
  \includegraphics[width=\linewidth]{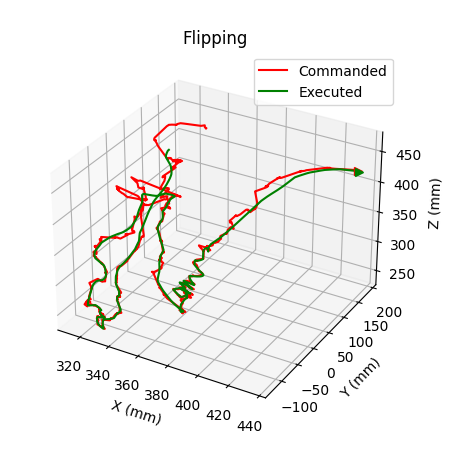}
  \caption{Flipping}
  \label{fig:5}
\end{subfigure}\hfil 
\begin{subfigure}{0.25\textwidth}
  \includegraphics[width=\linewidth]{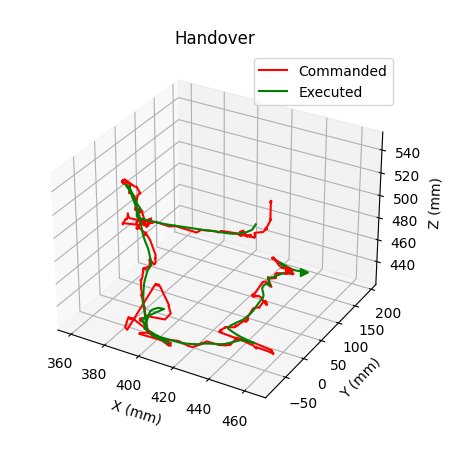}
  \caption{Handover}
  \label{fig:6}
\end{subfigure}\hfil 
\begin{subfigure}{0.25\textwidth}
  \includegraphics[width=\linewidth]{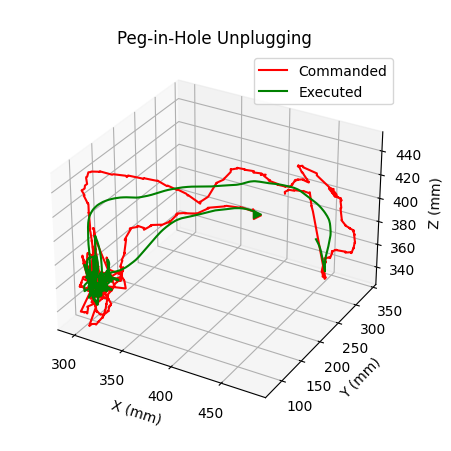}
  \caption{Peg-in-hole (Unplugging)}
  \label{fig:7}
\end{subfigure}\hfil
\begin{subfigure}{0.25\textwidth}
  \includegraphics[width=\linewidth]{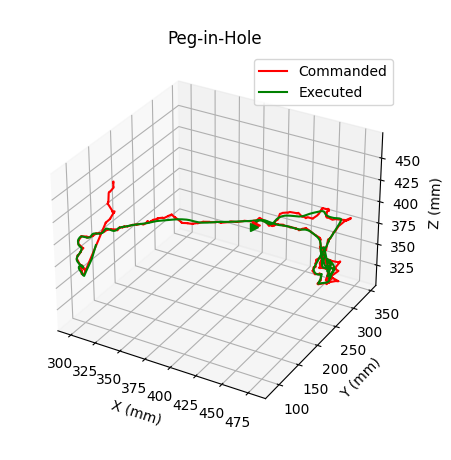}
  \caption{Peg-in-hole}
  \label{fig:8}
\end{subfigure}
\caption{Visualizations of commanded and executed 3D position trajectories from the different manipulation tasks. The triangular points indicate where the trajectory started.}
\label{fig:images}
\end{figure*}

While this problem may occur in any robot platform, it is especially prevalent in industrial robot platforms that have black-box controllers and built-in fault systems. We propose addressing this problem with a simplified filtering approach designed to accommodate the lack of certainty over the robot's dynamics and built-in fault system. More formally, we assume a commanded delta end-effector pose $\Delta$p is commanded at a given frequency $f$, and our goal is to produce a suitable filtered command signal $\Delta$p' for satisfying the command without faulting. We propose a two-stage approach, where we first transform $\Delta$p based on a speed cap and noise filtering threshold, and then use a linear transformation to interpolate from the current position to the desired end-effector pose at a higher frequency $f$' $>$ $f$. The first step ensures that the commanded end-effector pose does not activate the faults and that overshooting for the desired pose is mitigated, while the second step ensures that the robot is operating in a reactively to new commanded inputs from the user. Our algorithmic implementation for the two-stage filtering approach can be found in Algorithm \ref{algo}.

\begin{figure*}
\begin{center}
\begin{tabular}{||c c c c c||} 
 \hline
 Task & Time (s) & RMS (mm) & Comm. Jerk & Exe. Jerk \\ [0.5ex] 
 \hline\hline
 Bandu & 69.675 & 28.87 & 0.000611 &  0.000147 \\ 
 \hline
 Block destacking & 83.07 & 28.21 & 0.000617 & 0.000125 \\
 \hline
 Block stacking & 71.92 & 21.5 & 0.000525 & 0.000124 \\
 \hline
 Block stacking (irreg.) & 80.35 & 25.63 & 0.000632 & 0.000124\\
 \hline
 Flipping & 61.02 & 15.34  & 0.000432 & 0.000134 \\
 \hline
 Handover & 41.27 & 19.2  & 0.000602 & 0.000152 \\
 \hline
 Peg-in-hole unplugging & 85.62 &  38.382 & 0.000635  & 0.000134  \\
 \hline
 Peg-in-hole & 49.67 & 38.38 & 0.000542 & 0.000122  \\ [1ex] 
  \hline
\end{tabular}
\end{center}
\caption{A table of the task competition times, RMS between commanded and executed trajectories and jerk between the commanded and the executed trajectories.}
\label{table:time}
\end{figure*}

\section{EXPERIMENTS}
To evaluate our proposed filtering approach for improving the efficacy of VR teleoperation for industrial robot arms, we tested our ability to perform $8$ different complex manipulation tasks when using our proposed scheme on the Assista robot manipulator. Furthermore, we operate the robot in the stiffness control mode with a suitable choice of stiffness matrix for all the task to avoid excessive contact forces during manipulation.

For each task, we had an experienced operator (one of the author's of this paper) use the VR teleoperation system with our filtering approach implemented to perform each manipulation task. For every task, the robot was initially homed to the same starting configuration before recording. Each task was performed 2 times, and we recorded both the command trajectory from the user and the trajectory executed by the robot. Sample images of each of the tasks can be found in Figure \ref{fig:tasks1} and \ref{fig:tasks2}.

\subsection{Manipulation Tasks}
We evaluate the proposed VR teleoperation task on a variety of contact-rich manipulation tasks. Some of these tasks require precise placement and interaction during execution, and are challenging to solve using planning or Reinforcement Learning (RL)~\cite{9811812}. So, we try to understand how precise manipulation and interaction can be performed using the proposed system. In particular, we perform the following tasks using the proposed system:
\begin{enumerate}
    \item Bandu - Stack four Bandu blocks on top of each other in two columns (Figure \ref{fig:bandu}).
    \item Block stacking - Stack three colored cubes on top of each other into a column (Figure \ref{fig:block-stacking}).
    \item Block destacking - Destack three cubes and place them into a new stacked column (Figure \ref{fig:unstacking}).
    \item Block stacking irregular - Stack three irregular-shaped blocks on top of each other into a column (Figure \ref{fig:irregular}).
    \item Object flipping - Flip a bottle from a horizontal position to a vertical position using non-prehensile manipulation (Figure \ref{fig:flipping}).
    \item Handover - Pick up a screw driver from a slot and hand it over to a co-located human (Figure \ref{fig:handover}).
    \item Peg in hole - Lift a peg from a peg holder and place it into a fitted hole (Figure \ref{fig:peg-in-hole}).
    \item Peg in hole unplugging - Unplug a peg from a tight hole and put it back in a holder. (Figure \ref{fig:unplugging}).
\end{enumerate}

\section{Discussion}
Sampled visualizations of the commanded and executed trajectories for the different manipulation tasks can be found in Figure \ref{fig:images}. We also report the average task completion time for each of the manipulation tasks, the RMS between the commanded and executed trajectories, and the average jerk of the trajectories, which are included in Table \ref{table:time}. Qualitatively, we can see that there are a wide-range of different motion trajectories across the different tasks. Tasks like Block stacking and the handover task involve many linear motions associated with space motion, whereas tasks like Bandu and block destacking (irregular) involved a lot of subtle, small reposition movements due to the contact-rich nature involved in successfully completing the tasks. Whereas the commanded trajectories from the VR involved are relatively not smooth (due in part to operator error and position tracking error), our filtering approach results in a smoother trajectory that still keeps relatively low tracking error with the commanded signal. This is quantitatively evaluated based on the average jerk norm of the trajectories, which, as can be seen in Table \ref{table:time}, are much lower for the executed trajectories compared to the commanded trajectories. We found that this was crucial for tasks where motions with high jerk easily resulted in task failure, such as the flipping task.

\section{CONCLUSION}
We address the problem of using commercially-available VR hardware to effectively teleoperate an industrial robot manipulator. We propose a simplistic filtering scheme for processing position-commanded signals that reduces the number of faults and tracking error when teleoperating an industrial robot arm with a black-box controller. We demonstrate the efficacy of our approach by teleoperating a MELFA Assista robot to perform $8$ dexterous manipulation tasks. Future work will investigate leveraging our system and interface to enable industrial end-users to teach robot motor skills via imitation learning.
\bibliographystyle{IEEEtran}
\bibliography{root}

\end{document}